%% file: ms.tex
\documentclass[10pt,journal,compsoc]{IEEEtran}

\ifCLASSOPTIONcompsoc
  \usepackage[nocompress]{cite}
\else
  \usepackage{cite}
\fi

\ifCLASSINFOpdf
\usepackage[pdftex]{graphicx}
\else
\fi

\usepackage{amsmath}
\usepackage{array}
\usepackage{tabularx,multirow}
\usepackage[dvipsnames]{xcolor}
\usepackage{multirow}
\usepackage{siunitx}
\usepackage{mwe}
\usepackage{graphicx}
\usepackage{subcaption}
\usepackage{caption}

\begin{document}

\title{A study of the effect of the illumination model on the generation of synthetic training datasets}

\author{Xin Zhang, Ning Jia and Ioannis Ivrissimtzis 
\thanks{Xin Zhang, Ning Jia and Ioannis Ivrissimtzis are with the Department of Computer Science, Durham University}
\thanks{e-mail:\{xin.zhang3, ning.jia, ioannis.ivrissimtzis\}@durham.ac.uk}}

\markboth{Effect of illumination model on synthetic training data generation}
{}

\IEEEtitleabstractindextext{
\begin{abstract}
The use of computer generated images to train Deep Neural Networks is a viable alternative to real images when the latter are scarce or expensive. In this paper, we study how the illumination model used by the rendering software affects the quality of the generated images. We created eight training sets, each one with a different illumination model, and tested them on three different network architectures, ResNet, U-Net and a combined architecture developed by us. The test set consisted of photos of 3D printed objects produced from the same CAD models used to generate the training set. The effect of the other parameters of the rendering process, such as textures and camera position, was randomized. 

Our results show that the effect of the illumination model is important, comparable in significance to the network architecture. We also show that both light probes capturing natural environmental light, and modelled lighting environments, can give good results. In the case of light probes, we identified as two significant factors affecting performance the similarity between the light probe and the test environment, as well as the light probe’s resolution. Regarding modelled lighting environment, similarity with the test environment was again identified as a significant factor. 
\end{abstract}

}

\maketitle
\IEEEdisplaynontitleabstractindextext
\IEEEpeerreviewmaketitle

\hfill 
 
\input{sec1.tex}

\input{sec2.tex}

\input{sec3.tex}
\input{sec4.tex}

\input{sec5.tex}

\bibliographystyle{IEEEtran}
\bibliography{references}

\end{document}

%% file: sec1.tex
\section{Introduction}
\label{section I}

In the past few years deep Neural Networks (DNNs) have been firmly established as the state-of-the-art in a variety of fundamental computer vision tasks, such as segmentation, object recognition, or scene classification. The superiority of DNNs over classical machine learning models is more pronounced in data rich situations, where training with large annotated datasets allows millions of parameters to be tuned and stored in the network's deep hierarchical structures. When data is scarce, or when annotation requires high-level expertise and it is prohibitively expensive, various techniques have been developed allowing DNNs to deal with smaller data sets, such as, {\em transfer learning}, {\em data augmentation} and the use of {\em synthetic data}. 


The construction of synthetic training datasets in particular, has been successfully employed to facilitate the application of deep learning on several specialized computer vision tasks \cite{heimann2013learning,riegler2015anatomical,richardson20163d,mahmood2018deep}. There are also several more general studies, employing 3D modelling and rendering techniques at various levels of sophistication for synthetic image generation. On the higher end, advanced 3D graphics generators, such as UnrealStereo \cite{zhang2016unrealstereo}, have been used to generate photo-realistic renderings and their corresponding ground truths to train neural networks for optical flow\cite{li2018interiornet,tremblay2018falling}, semantic segmentation, and stereo estimation. At the lower end, addressing the limitation that sophisticated software requires designers at artist level, techniques such as domain randomization (DR)~\cite{tobin2017domain} have been proposed. In its original application, domain randomization has been used to train robots to recognize objects with simple shapes, essentially by forcing the network to focus on the main features of the objects and dispensing with photorealism.

In this paper, in the context of synthetic training dataset generation, we study the effect of the illumination model on DNN performance. To the best of our knowledge, this relationship has not been studied in the literature. In particular, there are no systematic comparisons of the performance of DNNs when different illumination models are used for generating a synthetic training dataset, indeed, in most cases, illumination model becomes a convenience choice dictated by the capabilities of the rendering software. In fact, quite often the model is not even reproducible, as it is not adequately described in the paper.

Here, in a systematic study of the influence of the illumination model, we compare the DNN performance over eight training datasets, generated with different illumination models. Five of them are {\em light probes}, i.e., models of outdoors or indoors natural lighting environments captured by specialized sensor systems at various resolutions. We also use three synthetic illumination models of increasing levels of complexity.

\begin{figure}
	\centering
	\includegraphics[width=0.45\columnwidth]{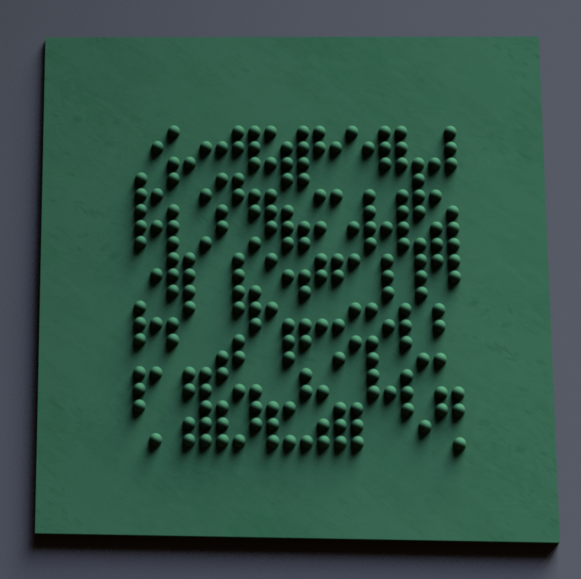} \hskip 0.5cm
	\includegraphics[width=0.45\columnwidth]{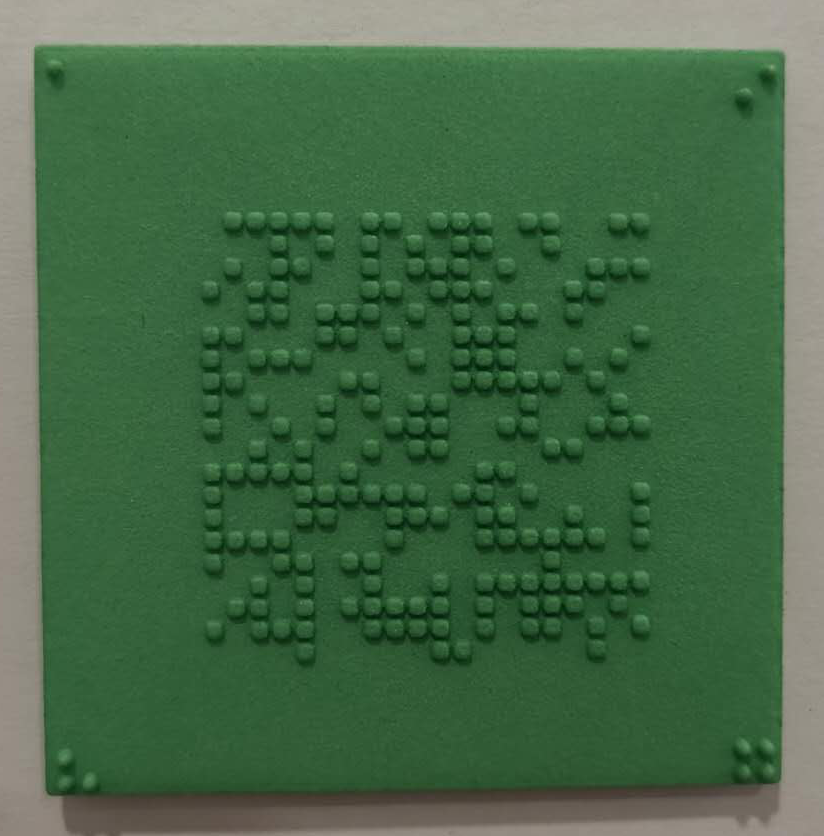} 	
	\caption{A synthetic training image (left1) and a test image (left2), which is the photo of a 3D printed object.} 
	\label{fig:trainingImage}
\end{figure}

The testbed for the comparison of the various illumination models is watermark retrieval from 3D printed objects \cite{zhang2018watermark}. The watermarks are embedded on flat surfaces of 3D printed objects and encode information in $20 \times 20$ bit arrays, where a bit value equal to 1 corresponds to a semi-spherical bump on the object's surface, Figure~\ref{fig:trainingImage}. In a first stage, the retrieval algorithm, which decodes the watermark from one or more photos of the physical 3D printed object, uses a DNN to generate a confidence map of the location of the bumps, and then uses a standard image processing pipeline to extract the bit-array from the confidence map. The overall performance of the retrieval algorithm largely depends on the performance of the DNN on two very basic computer vision task, the {\em detection} and the {\em localization} of the semi-spherical bumps. 

The use of 3D printed models for all testing aims at neutralizing the influence of the geometric modelling part of the graphics pipeline. Indeed, the same digital 3D models that are used for generating the synthetic training images are also used for 3D printing the physical objects, making the underlying 3D geometries in the training and the test sets almost identical. Since differences in the geometry between the training and test sets are expected to contribute significantly to the DNN error, their elimination reduces the risks to the validity of the results. To reduce the bias introduced by the  architecture of the DNN, we experimented with three different networks; ResNet50 \cite{he2016deep}, which gives low noise confidence maps, a variant of U-Net \cite{ronneberger2015u}, which gives  good localization, and a combination of the two networks we developed, which to some extent combines the strengths of the previous two networks.

\textbf{Contribution:} We present a systematic study of the relationship between the illumination model and the quality of  synthetic datasets generated for DNN training. The main findings can be summarized as follows: 
\begin{itemize}
\item The resolution of the light probe is a critical factor regarding its suitability for synthetic data generation. The similarity between the light probe and the illumination of the test environment is also important. 
\item Synthetic illumination models can give results that are competitive to those obtained from light probes. However, a good level of fidelity towards the test environment is required, as well as  
\end{itemize}
As a secondary contribution, we note that a characteristic of our approach, linked to the use of 3D printing technology, is that the the geometries of the digital objects in the training set and the physical objects in the test set are identical; they correspond to the same CAD models. In this novel setting, we are able to confirm and illustrate interesting characteristics of DNN behaviour, in particular, that ResNet50 generates confidence maps with lower noise, while U-Net outperforms ResNet50 in terms of localisation.

%% file: sec2.tex
\section{Background}

DNNs are the current state-of-the-art in various fundamental computer vision problems such as, object detection~\cite{ren2015faster,liu2016ssd,dai2016r,redmon2016you,he2017mask} and semantic segmentation~\cite{dai2016instance,li2017fully}. 

The various extensions of Convolutional Neural Networks (CNNs), such as Fast R-CNNs, R-FCNs and SSDs are widely considered as some of the best performing networks for object detection. In \cite{girshick2015fast}, the use of a classical model, Fast R-CNN, was proposed for deep learning-based object detection. The Fast R-CNN is an end-to-end training object detection framework that uses multi-task loss on each labelled Region of Interest (RoI) to jointly train for classification and bounding-box regression. Moreover, unlike SSPnet~\cite{he2015spatial}, Fast R-CNN training can update all network layers. In \cite{dai2016r}, a region-based fully convolutional network (R-FCN) was proposed for objects detection. The R-FCN considers each {\bf region proposal}, divides it up into sub-regions and iterates over those sub-regions. In this process, the R-FCN is able to use each generated position-sensitive score map to encode the position information with a relative spatial position. R-FCNs have higher accuracy and are several times faster than Fast R-CNNs. In \cite{liu2016ssd}, a single-shot detector (SSD) was proposed, processing the images to simultaneously estimate bounding boxes for the detected objects and predictions for their class, that is, the SSD does not only use the network to generate RoIs, but simultaneously classifies those regions. 

Other popular neural network models tend to be fairly similar to the three described above, and they mostly rely on very deep CNN's architectures, such as ResNet~\cite{he2016deep}. In~\cite{he2016deep}, the deep residual learning technique was introduced and ResNet was proposed as a first example implementing that architecture. Its main novelty was the introduction of residual links, shortcut connections enabling cross-layer connectivity, facilitating the convergence of the training process by avoiding gradient diminishing problems. To this day, ResNet remains one of the most powerful network architectures available. 

U-Net was proposed in \cite{ronneberger2015u} as an alternative to very deep CNN architectures such as ResNet. It is a lighter network sharing many common features with FCNs. The main difference between FCNs and U-Net, is the latter's symmetric architecture, which concatenates the feature maps in the expanding path with the corresponding cropped feature maps in the contracting path, while, in contrast, in FCNs feature maps are summed. 

In~\cite{he2016deep} it was shown that ResNet has a very large receptive field, which means that the network retrieves more coarse information and can ignore tiny details. In contrast, due to the skip connections, U-Net is good at retrieving information corresponding to the image details~\cite{ronneberger2015u}, making it a popular choice in the medical domain. As illustrated in~\cite{he2019semi} on a typical skin analysis application, the outputs of the U-Net show more details and give higher IOU rates than those of ResNet50, but also contain more noise.

\subsection{Synthetic dataset generation}

Synthetic dataset generation is an efficient alternative to the use of natural images in DNN training, especially when natural images are scarce, or expensive to obtain, or their annotation requires extensive expert input and is thus costly. On the other hand, synthetic data generation might have a quite steep overhead in the construction of the 3D models, but then the generation of abundant synthetic data becomes a quite straightforward and low-cost process. Moreover, in various application scenarios, annotations serving as ground truths not only can be automatically created at a minimal additional cost, but most importantly, they can be exact, free of any human introduced errors. 

Synthetic data have been widely used in training deep learning networks. In \cite{jaderberg2014synthetic} and~\cite{gupta2016synthetic}, synthetic datasets are generated by pasting text images are on black and various natural backgrounds, respectively. In \cite{dosovitskiy2015flownet},  renderings of 3D chair images are combined with natural image backgrounds to train FlowNet. In \cite{handa2016understanding}, indoors scenes are built using CAD models to train a network for indoors scene understanding. In \cite{atapour2018real}, a depth estimation model is trained using synthetic scene data generated by a game engine. In \cite{zhang2016unrealstereo}, synthetic images train a network for stereo vision understanding. The first three of the above methods used existing image datasets and 3D models, while the last two utilised purpose built artist level modelling.

Domain randomization is a simpler technique for training networks on simulated images. It randomizes some of the parameters of the simulator and generates a dataset of sufficient variety consisting of non-artistic synthetic data~\cite{tobin2017domain,terry2012additive}. Here, we borrow from the domain randomisation technique in the way we treat some other important rendering variables that are not directly related to the illumination model, specifically, texture and camera position and orientation.

%% file: sec3.tex
\section{Experimental setup}
\label{sec:setup}

In this section, we first describe the test dataset, consisting of images of 3D printed objects captured by a mobile camera. Next, we describe the generation of the synthetic training datasets, focusing on the illumination model, which is the experimental variable we study here. Finally we describe the architecture of the neural networks of the experiment. 

\subsection{Test dataset}

\begin{figure}
	\centering
	\includegraphics[width=0.24\columnwidth, height=2.3cm]{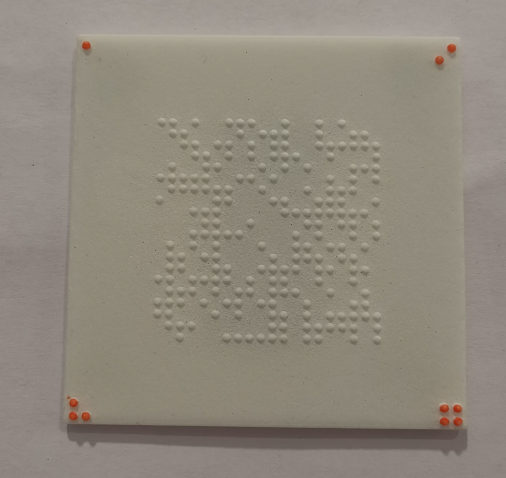} \hfill 
	\includegraphics[width=0.24\columnwidth, height=2.3cm]{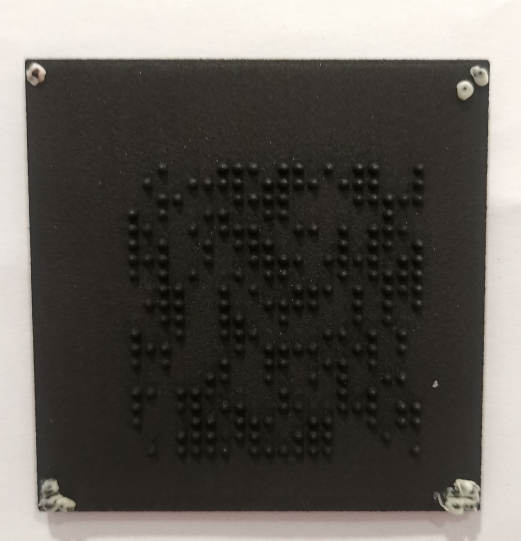} \hfill
	\includegraphics[width=0.24\columnwidth, height=2.3cm]{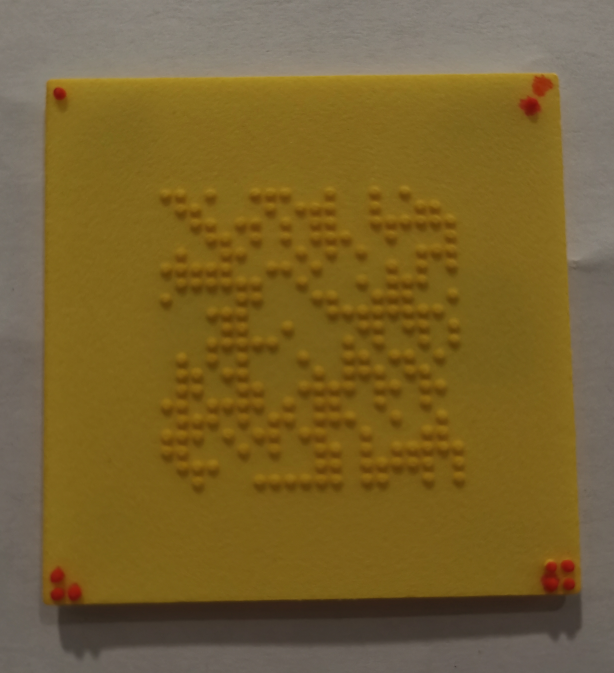} \hfill 
	\includegraphics[width=0.24\columnwidth, height=2.3cm]{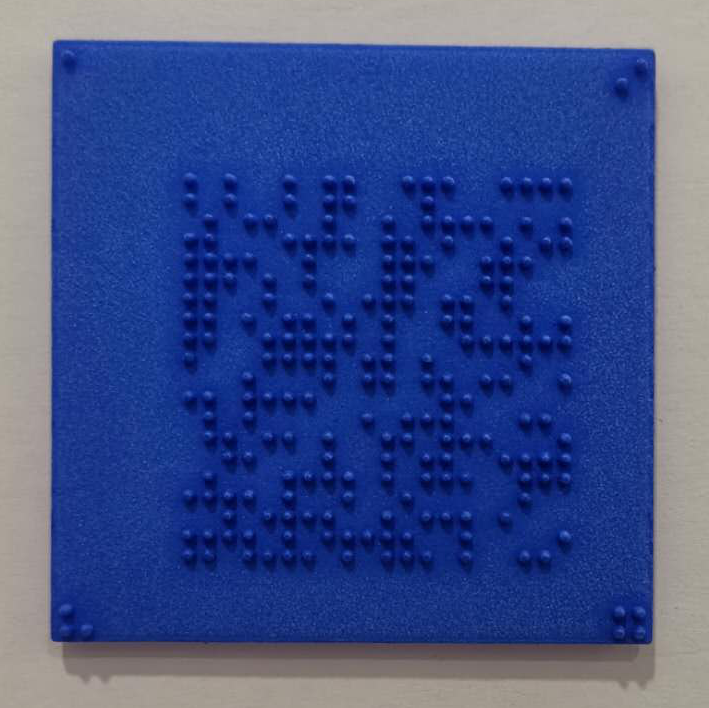} 
	\includegraphics[width=0.24\columnwidth, height=2.3cm]{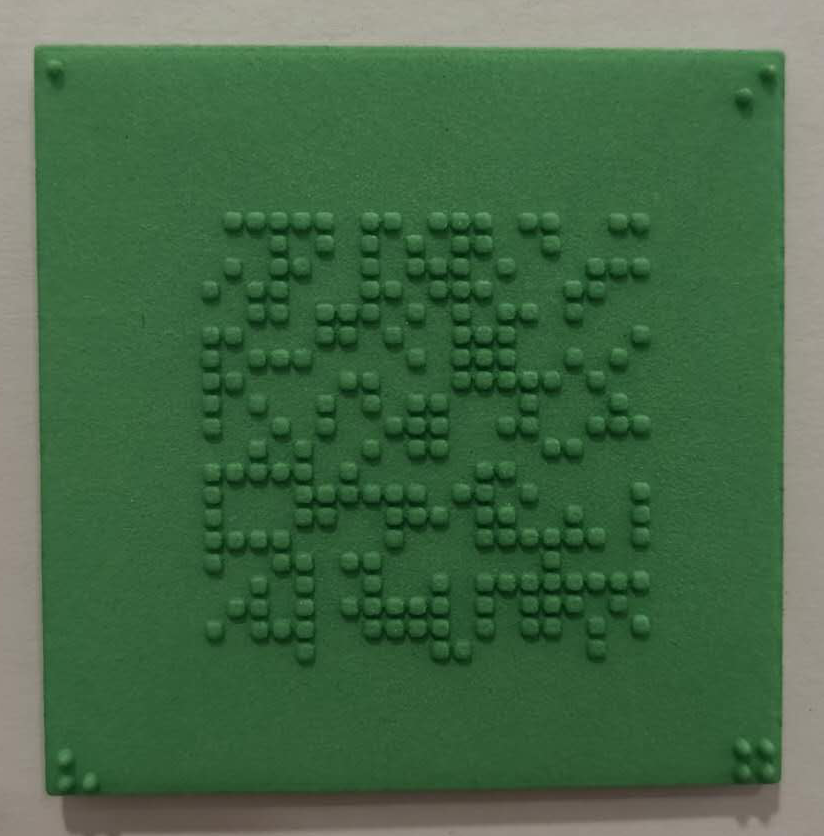} \hfill 
	\includegraphics[width=0.24\columnwidth, height=2.3cm]{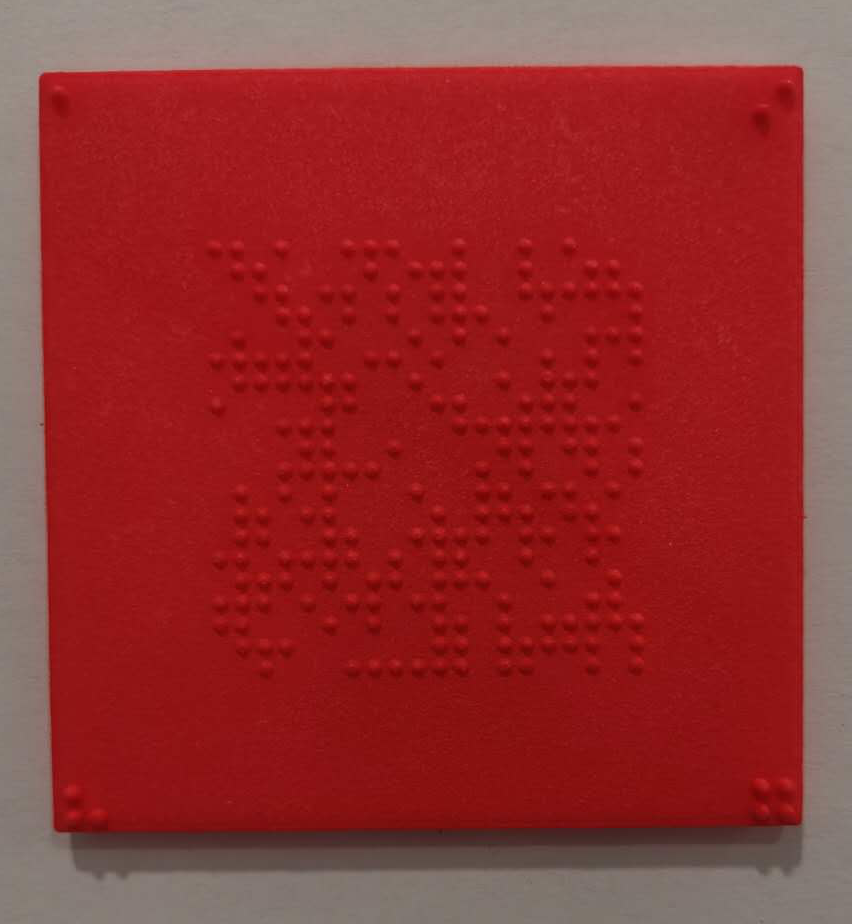} \hfill
	\includegraphics[width=0.24\columnwidth, height=2.3cm]{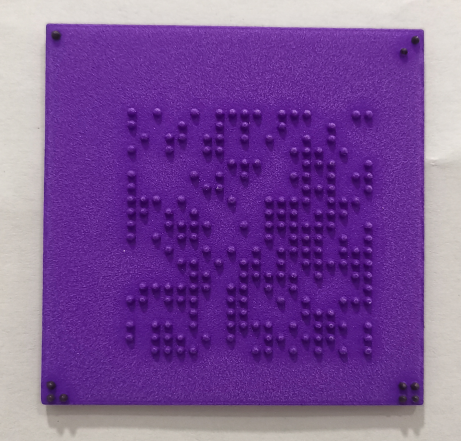} \hfill 
	\includegraphics[width=0.24\columnwidth, height=2.3cm]{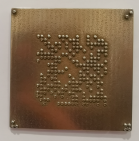} 
	\caption{The 3D printed objects.}
	\label{fig:3dprintedModels}
\end{figure}

We 3D printed eight watermarked objects in total, seven of them on plastic of various dark and bright colours and one in metal. Each object was printed on a single material and colour. The watermarks are $20 \times 20$ arrays of randomly generated bits, encoded by semi-spherical bumps, the size of which varies from object to object. Each printed object carries a different watermark to ensure that high watermark retrieval accuracy means that the network can successfully detect and localise bumps on the object's surface, rather than memorize their relative locations. Figure~\ref{fig:3dprintedModels} shows the eight printed watermarked objects. 

\begin{figure}
	\centering
	\includegraphics[width=0.45\columnwidth, height=3cm]{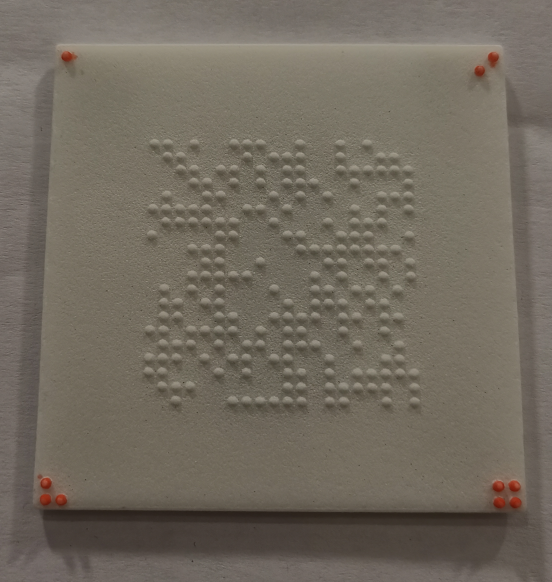} \hfill 
	\includegraphics[width=0.45\columnwidth, height=3cm]{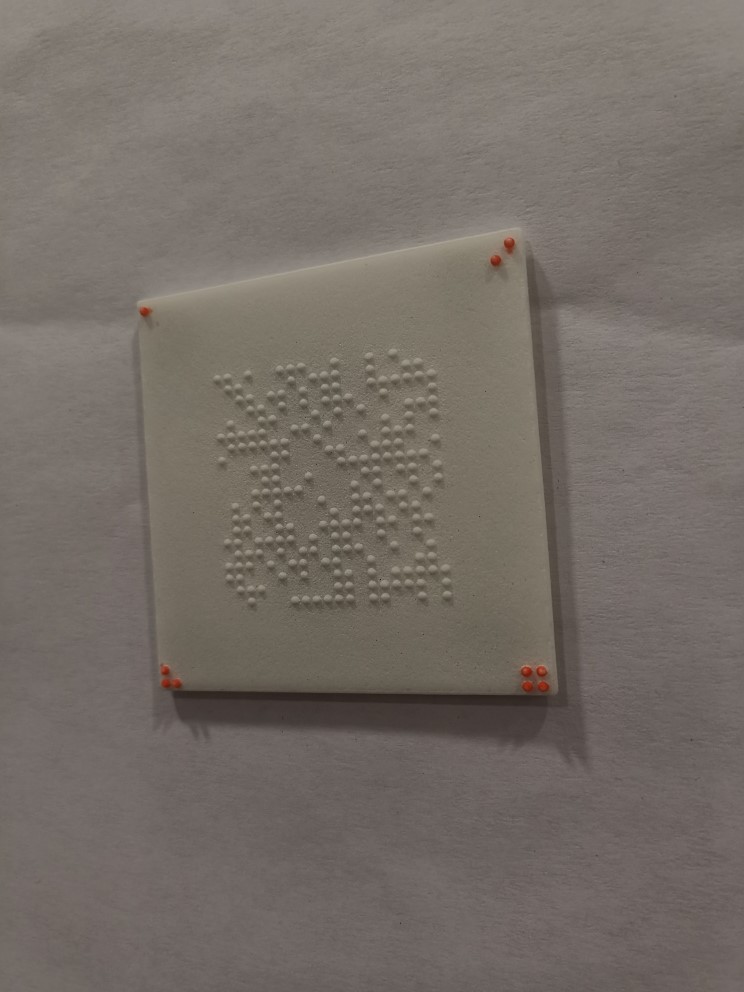} 
	\caption{Test images from a high and a low vertical angle.}
	\label{fig:cameraPosition}
\end{figure}

The test dataset consisted of 20 images from each object, 160 in total, which means that each neural network / training dataset combination was tested on the detection and localisation of $160\times 400 = 64,000$ possible bumps. The images were taken indoors, under natural light coming from windows and artificial light from the ceiling. The background was monochrome and had a fine texture, ensuring that any background textural features were smaller than the watermark bumps. The images were shot from various camera positions: the distance was roughly 1m, the horizontal angle varied from $0^\circ$ to $360^\circ$, and the vertical angles, expected to have the most significant effect on the performance of the retrieval algorithms, varied from $90^\circ$ (looking at the object from the top) to about $45\circ$. The camera was pointing to the target 3D printed object, with small errors introduced in purpose. Figure~\ref{fig:cameraPosition} shows two test images, one from a large and one from a small vertical angle.

\subsection{Synthetic Image generator}

The training dataset consists of rendered images of the CAD models of the 3D printed objects. We note that the levels of noise introduced by the current, consumer level 3D printing processes, can be significant, and part of the error in the watermark retrieval can be attributed to imperfections of the 3D printing process. Nevertheless, by choosing a commercial 3D printing service, rather than in-house printing, and by sending to the printers the exact 3D models that were used in the generation of the training sets, we neutralized the effect of 3D geometry as much as possible.

For all rendering we used Mitsuba~\cite{mitsuba}, which, being an open source software with over 100 plug-ins ranging from textures and materials to light sources, supported all the variables of our experimental setup as described below. 

\textbf{Illumination model:} The illumination model used in the generation of the training dataset is the main variable of study in the experiment. Table~\ref{table:illuminationModels} lists the eight illumination models that were used to create the corresponding eight training datasets. 
\begin{table}
	\centering
	\caption{Illumination models.}
	\begin{tabular}{|c|l|c|}
		\hline
		 & \multicolumn{1}{c|}{Illumination model} & Resolution \\
		\hline
		1 & Indoor dining room &  $3072 \times 6144$ \\ 
		2 & Indoor cathedral  & $1536 \times 3072$ \\		
		3 & Outdoor glacier & $1024 \times 2048$ \\
		4 & Indoor kitchen & $640 \times 640$ \\
		5 & Outdoor, covered hallway & $640 \times 640$ \\ \hline
		6 & Sky, physically based skylight at 10am & synthetic \\
		7 & Indoor office, area lights & synthetic \\
		8 & Indoor classroom, area lights and skylight & synthetic \\
		\hline
	\end{tabular}	
	\label{table:illuminationModels}
\end{table}

\begin{figure*}
	\centering
	\begin{subfigure}{0.24\linewidth}
		\centering
		\includegraphics[width=\linewidth,height=2cm]{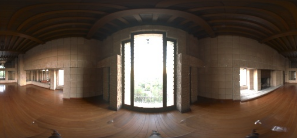}
		\caption{Indoor dinning room}
	\end{subfigure}
	\hfill
	\begin{subfigure}{0.24\linewidth}
		\centering
		\includegraphics[width=\linewidth,height=2cm]{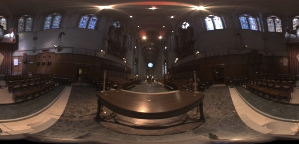}
		\caption{Indoor cathedral}	
	\end{subfigure}
	\hfill	
	\begin{subfigure}{0.24\linewidth}
		\centering
		\includegraphics[width=\linewidth,height=2cm]{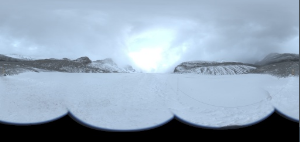}
		\caption{Outdoor glacier}
	\end{subfigure}
	\hfill	
	\begin{subfigure}{0.24\linewidth}
		\centering
		\includegraphics[width=\linewidth,height=2cm]{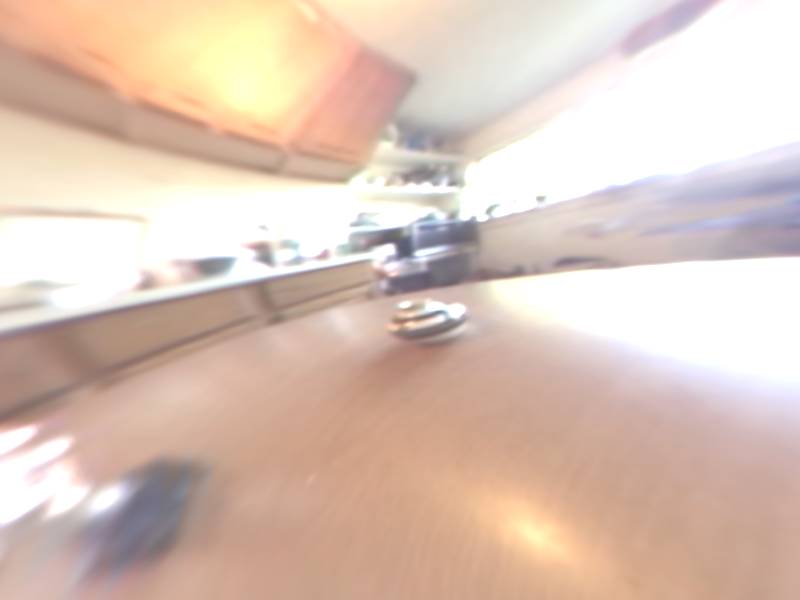}
		\caption{Indoor kitchen}
	\end{subfigure}
	\begin{subfigure}{0.24\linewidth}
		\centering
		\includegraphics[width=\linewidth,height=2cm]{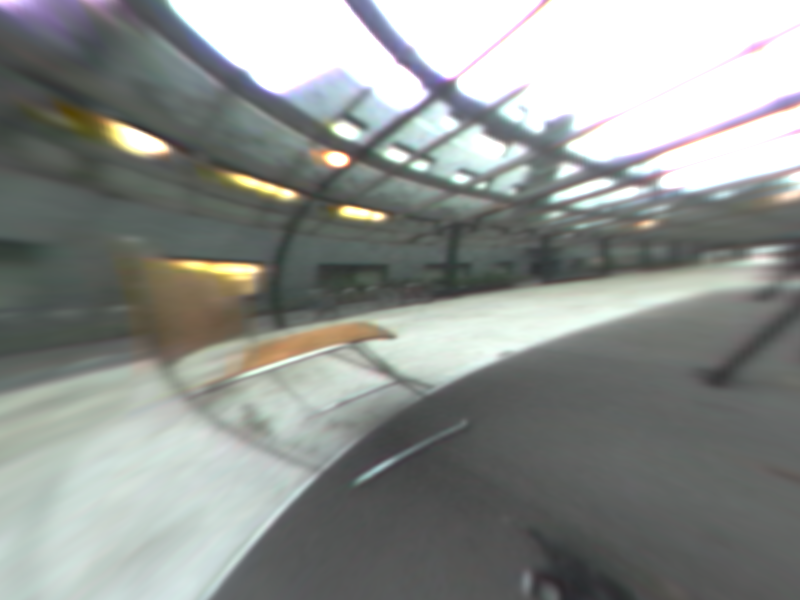}
		\caption{Outdoor, covered hallway}
	\end{subfigure}
	\hfill
	\begin{subfigure}{0.24\linewidth}
		\centering
		\includegraphics[width=\linewidth,height=2cm]{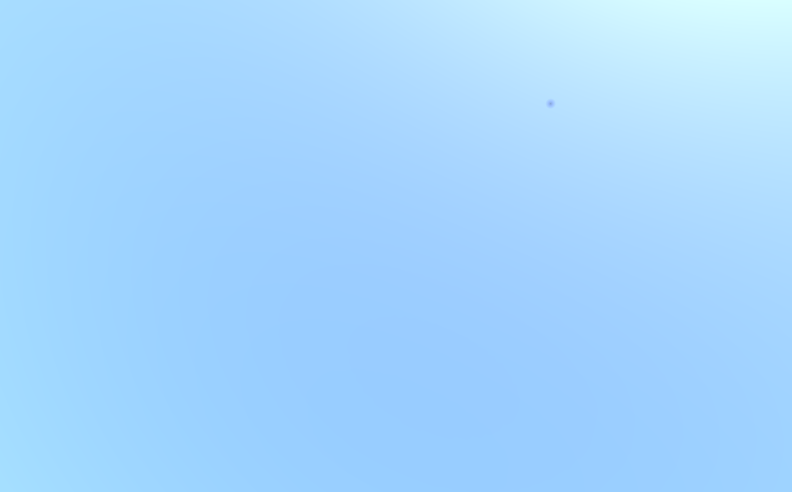}
		\caption{Sky, at 10am}
	\end{subfigure}
	\hfill
	\begin{subfigure}{0.24\linewidth}
		\centering
		\includegraphics[width=\linewidth,height=2cm]{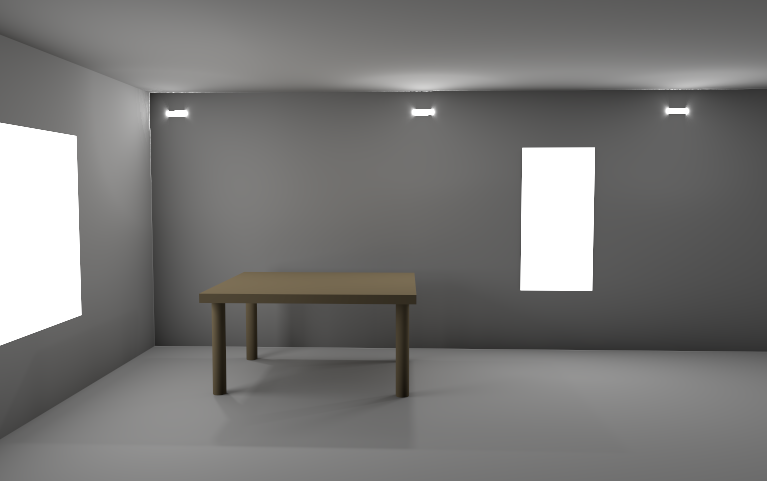}
		\caption{Indoor office}
	\end{subfigure}
	\hfill
	\begin{subfigure}{0.24\linewidth}
	\centering
	\includegraphics[width=\linewidth,height=2cm]{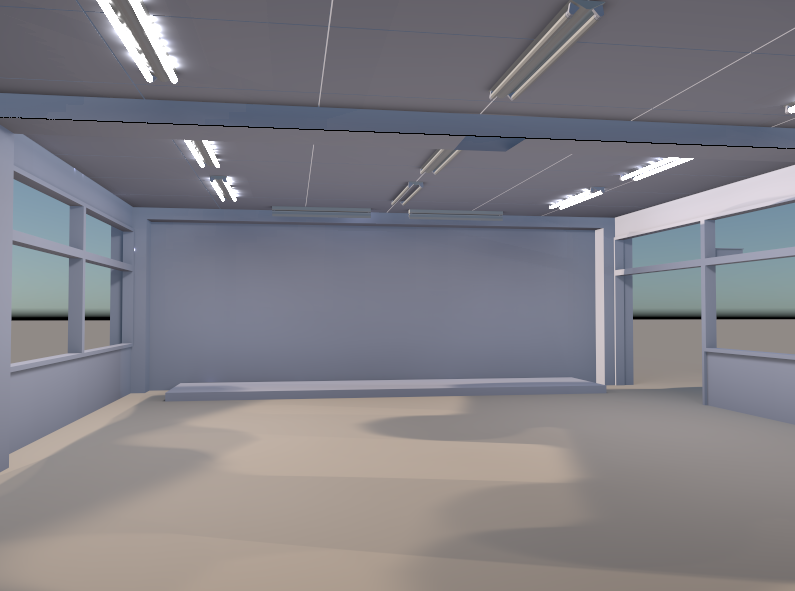}
	\caption{Indoor classroom}
\end{subfigure}
	\caption{The illumination models of the training sets.}
	\label{fig:illumination}
\end{figure*}

The first five models are based on light probes, i.e. images recording the incident light at a particular point in space. Figure~\ref{fig:illumination} (a-e) shows the five light probes we used, downloaded from~\cite{PaulDebevec} and~\cite{BernhardVogl}. They are at various resolutions, some capture outdoor and other indoor environments, and have various levels of similarity with the illumination conditions under which the test set was acquired. In particular, that similarity is high in the indoor dining room (\#1), indoor kitchen (\#4), and to some extend the covered hallway (\#5), while the cathedral (\#2) and the glacier (\#3) have their one distinct illumination models.

The three synthetic illumination models are shown in Figure~\ref{fig:illumination} (f-h). The first, is simulated 10am skylight generated by the physically based method proposed in \cite{preetham99}, and the scene is void of any geometry. The second, model is the model of an office room with large area lights simulating windows and some secondary wall lights, which nevertheless do not correspond to a feature of the test dataset environment. Finally, the third illumination model consists of a faithful 3D model of the test room, faithful ceiling lights, and 10am skylight entering from transparent windows. 

\begin{figure}
	\centering
	\begin{subfigure}{0.45\linewidth}
		\centering
		\includegraphics[width=0.5\textwidth]{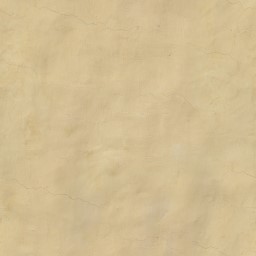}\hfill
		\includegraphics[width=0.5\textwidth]{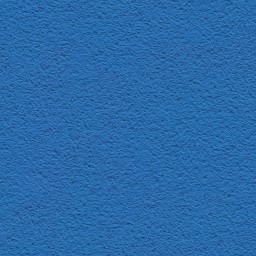}
		\includegraphics[width=0.5\textwidth]{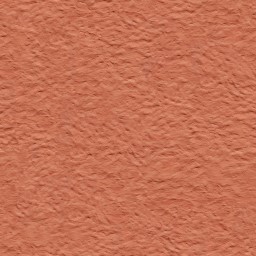}\hfill
		\includegraphics[width=0.5\textwidth]{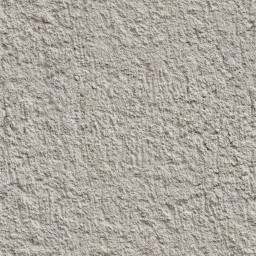}
		\caption{Texture map samples}
	\end{subfigure}
\hfill
	\begin{subfigure}{0.45\linewidth}
		\centering
		\includegraphics[width=\textwidth]{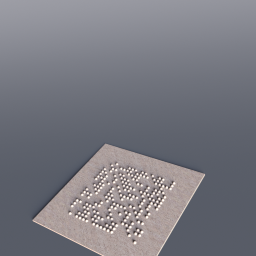}
		\caption{Synthetic image sample}
	\end{subfigure}
\caption{Texture map samples and training dataset image sample.}
\label{fig:texture}
\end{figure}

\textbf{Textures:} To simulate the colour and material diversity of the test objects, the generation of the training dataset is based on a wide range of colours and textures. The diffuse map is built upon the bitmap with additional diffuse reflectance, using the trilinear filter. To account for the variety of colors and materials in the 3D printed objects, we built a texture library containing 100 different texture images found using an online search engine. In addition, a monochrome planar background, which is rendered in gradient colour, is placed behind the watermarked 3D model to reduce the effect of the reflections.

\textbf{Camera position:} For each image the camera position is defined by a triplet of random numbers $(d, \theta, \phi)$. The camera is placed at a distance $d\in[50cm, 100cm]$ above the plane of the object, looks at the object from a vertical angle $\theta \in [45^\circ, 90^\circ]$, and revolves around the object by $\phi \in [0^\circ, 360^\circ$.


\subsection{Watermark detector}
\begin{figure*}
	\centering
	\includegraphics[width=0.75\textwidth]{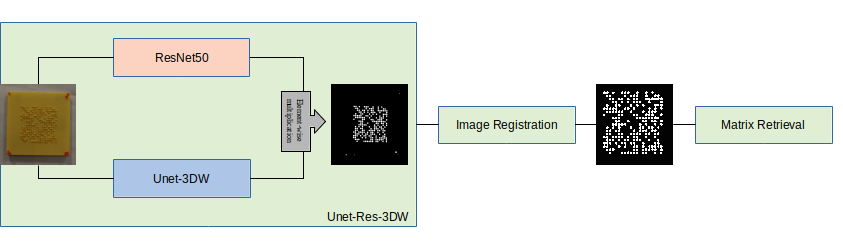}
	\caption{The flow diagram of the combined neural network.}
	\label{fig:flow}
\end{figure*}

The first component of the watermark retrieval algorithm is a neural network returning a confidence map encoding the probability a pixel is part of a bump, rather than background. We used an adaptation of U-Net network, which we called \textbf{Unet-3DW}, ResNet50, and a combination of the two. The subsequent components of the algorithm are an image registration module followed by a module that processes the registered confidence maps to retrieve the watermark bit matrix. Figure~\ref{fig:flow} shows the flow diagram for the case of the combined neural network. 

\subsubsection{DNN architectures}

\begin{figure}
	\centering
	\includegraphics[width=\linewidth]{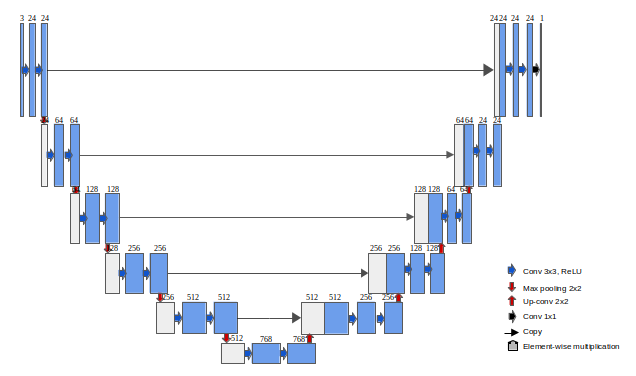}
	\caption{The Unet-3DW architecture.}
	\label{fig:unet}
\end{figure}

The standard U-Net model has four downsampling and four upsampling steps, each one halving or doubling the number of feature channels, respectively. In contrast, our \textbf{Unet-3DW} architecture consists of six downsampling and six upsampling steps, again laving and doubling the number of feature channels. In the contraction, each layer contains two repeated \(3\times3\) padded convolutions and a \(2\times2\) max pooling with stride \(2\), where each convolution is followed by a batch normalization (BatchNorm2d) and a rectified linear unit (ReLU). The central bottleneck only contains one convolutional layer. In the expanding part, each layer consists of two padded \(3\times3\) convolutions, each one followed by a ReLU. The first convolution concatenates two parts, the first comes from the corresponding feature map of the contraction path; the second is a \(2\times2\) up-convolution of the feature map. The second is a \(3\times3\) convolution. At the very final layer, each 24 feature vector is mapped to the desired confidence map by a \(1\times1\) convolution. 

The second network we employed is the \textbf{ResNet50}. The third DNN we employ, combines and balances ResNet and U-Net, see Figure~\ref{fig:flow}. As stated, for example, in~\cite{CNNsee}, the deep layers of a CNN learn high-level global features, which contain the most information about global patterns in the images, while low layers extract more local features. The combined network learns both global and local features, generating a confidence map by element-wise multiplication of the ResNet50 and Unet-3DW outputs. This is fed to the last layer with a LogSigmoid activation, which generates the final confidence map of the combined DNN. 


\subsubsection{DNN implementations}

We implemented both ResNet and  UNet-3DW in Pytorch and adopted the $BCEWithLogitsLoss$ for computing their loss.  $BCEWithLogitsLoss$ combines in a single class a sigmoid layer and the $BCELoss$, and is numerically more stable than using $BCELoss$ only. 
\begin{equation*}
	\begin{split}
	BECWithLogitsLoss(x,y)= \qquad \qquad \qquad \qquad \qquad \ \\
	\frac{1}{N}\sum_{i=1}^{N}[y_{i}\times\log(\sigma(x_{i}))
	+(1-y_{i})\times\log(1-\sigma(1-x_{i}))]
	\end{split}
\end{equation*}
where \(x\) is the predicted value, \(y\) is the target value and \(N\) is the total number of pixels.

The combined model is trained end-to-end with a joint loss function \(L\):
\begin{equation*}
L \quad=\quad \alpha \times L_{Unet-3DW}+\beta \times L_{ResNet} 
\end{equation*}
where \(\alpha\) and \(\beta\) are used to balance the weight of the two branches, and we found that \(\alpha = 1\) and \(\beta = 0.8\) works well for the problem at hand. 

The networks are optimised using Adam \cite{kingma2014adam}, the initial learning rate is \(1e -3\) and is decreased by \(1e-1\) after each 100 epochs until \(1e-5\). We used a single NVIDIA GeForce GTX TITAN X graphics card and trained our network with a batch size of 4. 

Because our training data are inexpensive synthetic images with controllable parameters, data augmentation such as mirroring, translational shift, rotation and relighting is not necessary during the training phase. Instead, for time and resource efficiency, we render our synthetic image with \(1024\times1024\) size, which is lower than the \(3072\times2048\) of the test images. 


\subsubsection{Image registration and watermark matrix retrieval}

The extraction of the embedded watermark matrix from the estimated confidence map is still a non-trivial task.

\begin{figure}
	\centering
	\includegraphics[width=2.8cm, height=2.8cm]{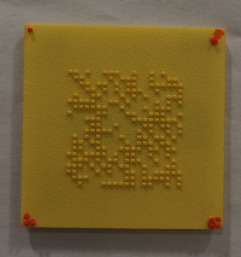} \hfill 
	\includegraphics[width=2.8cm, height=2.8cm]{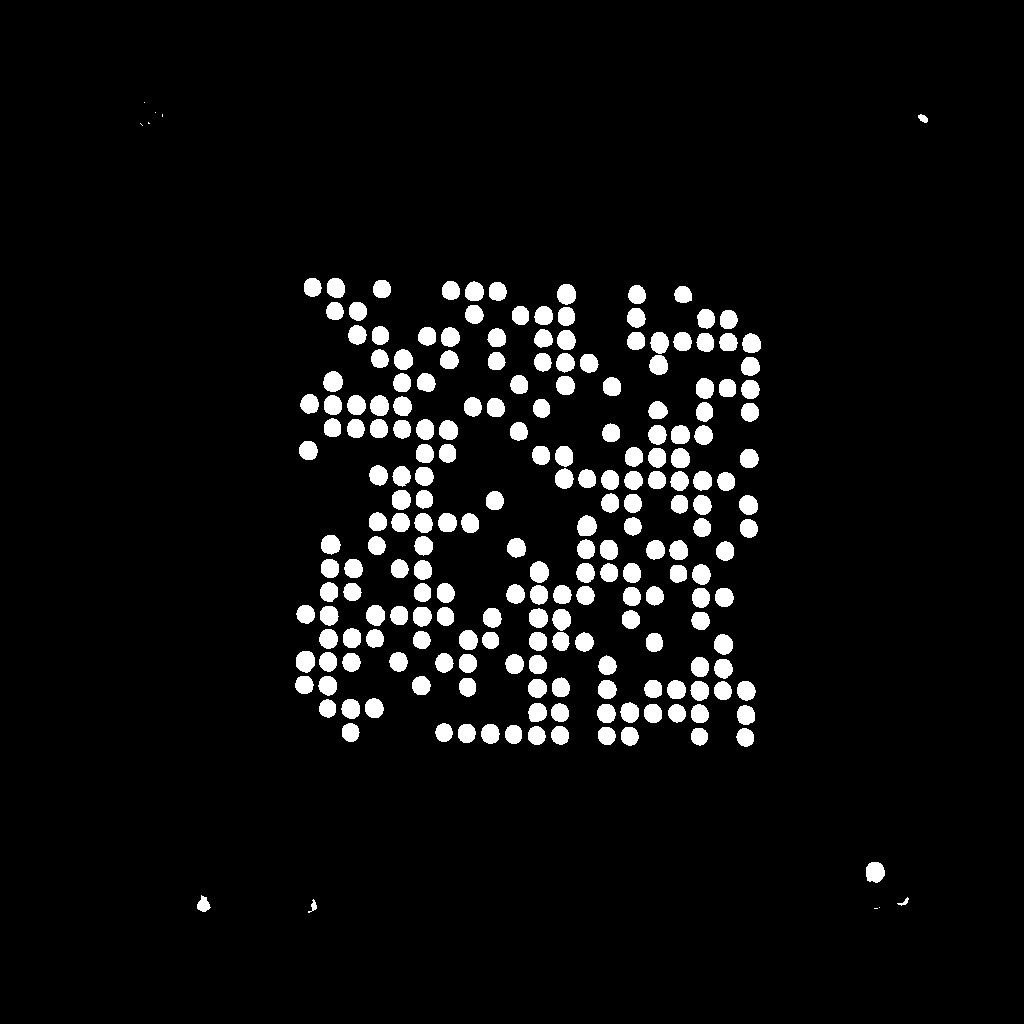} \hfill 
	\includegraphics[width=2.8cm, height=2.8cm]{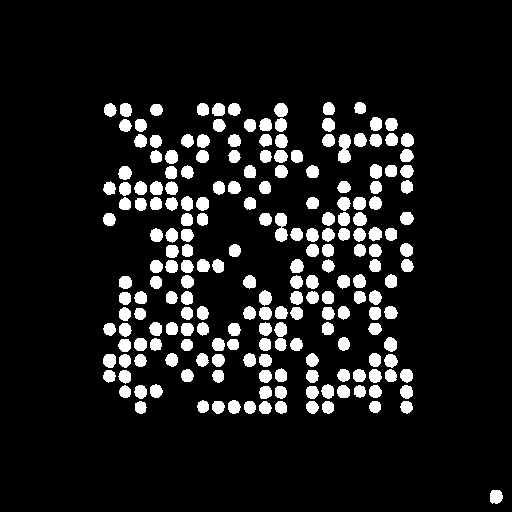} \hfill 
	\caption{Test image (left1), confidence map outputted by CNN-3DW (left2), binary confidence map obtained by regularisation followed by thresholding (left3).}
	\label{fig:seventh}
\end{figure}

Since a  significant error in the localisation of the watermarked region would lead to catastrophic error in the following stages, we decided to tackle this problem by annotating four \textit{landmarks} at the corners of the watermark region with a distinguishable colour (see Figure~\ref{fig:seventh}). During retrieval, the watermark regions are easily located by finding the differently colored landmarks at the corners, and use them to transform the quadrilateral watermark region into a square region (see Figure \ref{fig:seventh}). The registered confidence map is then binarized by Otsu thresholding, and we use Matlab's {\em regionprops} function to detect its connected regions, and obtain estimates of their centroids and their two semi-axes. If the sum of the two semi-axes is above a threshold, a bit value 1 is assigned to the centroid of that region, see Figure~\ref{fig:seventh}. 

As a final step, we need to extract the watermark matrix $M$ from the coordinates of the region-props centroids that have been assigned bit values 1. Ideally, there would exist only $m$ (the number of rows and columns of $\mathbf{M}$) unique values for the $x$ and $y$ coordinates, however, due to imperfections in the previous steps, the coordinates have biases. Instead, we use $K$-means clustering on the $x$ and the $y$ coordinates of the centroids, respectively. We set the number of clusters as $m$ and rank the $m$ cluster centres. The point falling into the $i$-th and $j$-th clusters, respectively, will correspond to the $(i,j)$ entry of the matrix $\mathbf{M}$. More details on this last part of the algorithm can be found in~\cite{zhang19}. 

%% file: sec4.tex
\section{Experiment}

To assess the effect of the illumination model on the quality of the generated synthetic data, we conducted a series of experiments on several network / training dataset combinations.  


\subsection{Training datasets and evaluation metrics}

We generated eight training datasets of 2K images each, one for each illumination model. Each image carries a distinct, randomly generated,  $20 \times 20$ watermark, and a randomly chosen texture from the set of 100 different textures, as described in section~\ref{sec:setup}. The camera positions are unique within each dataset, that is, images are captured under 2K different camera positions, but the same set of camera positions is used in all eight datasets. 


The performance metrics we report are the recall TPR (True Positive Rate) =TP/(TP+FN), precision PPV (Positive Predictive Value) = TP/(TP+FP) and F1 scores 2$\times$PPV$\times$TPR/(PPV+TPR), where TP and FN are the numbers of bits with value 1 retrieved correctly and correspondingly incorrectly, while FP is the number of bits with value 0 retrieved incorrectly. 

\subsection{Results}

In section~\ref{sec:networkAssessment}, we evaluate the watermark retrieval capabilities of the three networks we tested, aiming at selecting an effective retrieval algorithm. In section~ref{sec:illuminationModels}, we selected algorithm is used to explore the effect of the illumination model on the quality of the generated synthetic datasets.

\subsubsection{Assessing DNN performance}
\label{sec:networkAssessment}

We present results for each of the 8 test sets separately, each test set consisting of photos of a particulal 3D printed object. On each test set we tested $3\times 8 = 24$ network-training set combinations, and for each of the three networks we report the average F1 score over the 8 training sets. Table~\ref{table:f1scoresNetworks} summarizes these average F1 scores for Unet-3DW, ResNet50, and the combined Unet-Res-3DW. Note that all training data are synthetic images and all test data are real photos of 3D printed objects. 

\begin{table}[h]
	\centering
	\caption{Average F1-scores for Unet-3DW,  ResNet50 and Unet-Res-3DW.}
	\begin{tabular}{lccc}
		\hline
		 & {Unet-3DW} & {ResNet50} & {Unet-Res-3DW} \\
		\hline	
		red & 0.78 & 0.76 & {\bf 0.81} \\	
		yellow & 0.84 & 0.79 & {\bf 0.89} \\
		green & 0.83 & 0.82 & {\bf 0.90} \\
		white & {\bf 0.88} & 0.84 & {\bf 0.88} \\
		black & 0.77 & 0.72 & {\bf 0.83} \\
		blue & 0.77 & 0.79 & {\bf 0.83} \\
		purple & 0.72 & 0.74 & {\bf 0.75} \\
		metal & 0.76 & 0.73 & {\bf 0.80} \\
		\hline
	\end{tabular}
	\label{table:f1scoresNetworks}
\end{table} 

\begin{figure}
	\centering
	\begin{subfigure}[b]{0.45\columnwidth}
		\centering
		\includegraphics[width=4cm, height=4cm]{y1.png}
		\caption{Test image} 
		
	\end{subfigure}
	\hfill
	\begin{subfigure}[b]{0.45\columnwidth}  
		\centering 
		\includegraphics[width=4cm, height=4cm]{ycom.png}
		\caption{Unet-Res-3DW density map}
		
	\end{subfigure}
	\vskip\baselineskip
	\begin{subfigure}[b]{0.45\columnwidth}   
		\centering 
		\includegraphics[width=4cm, height=4cm]{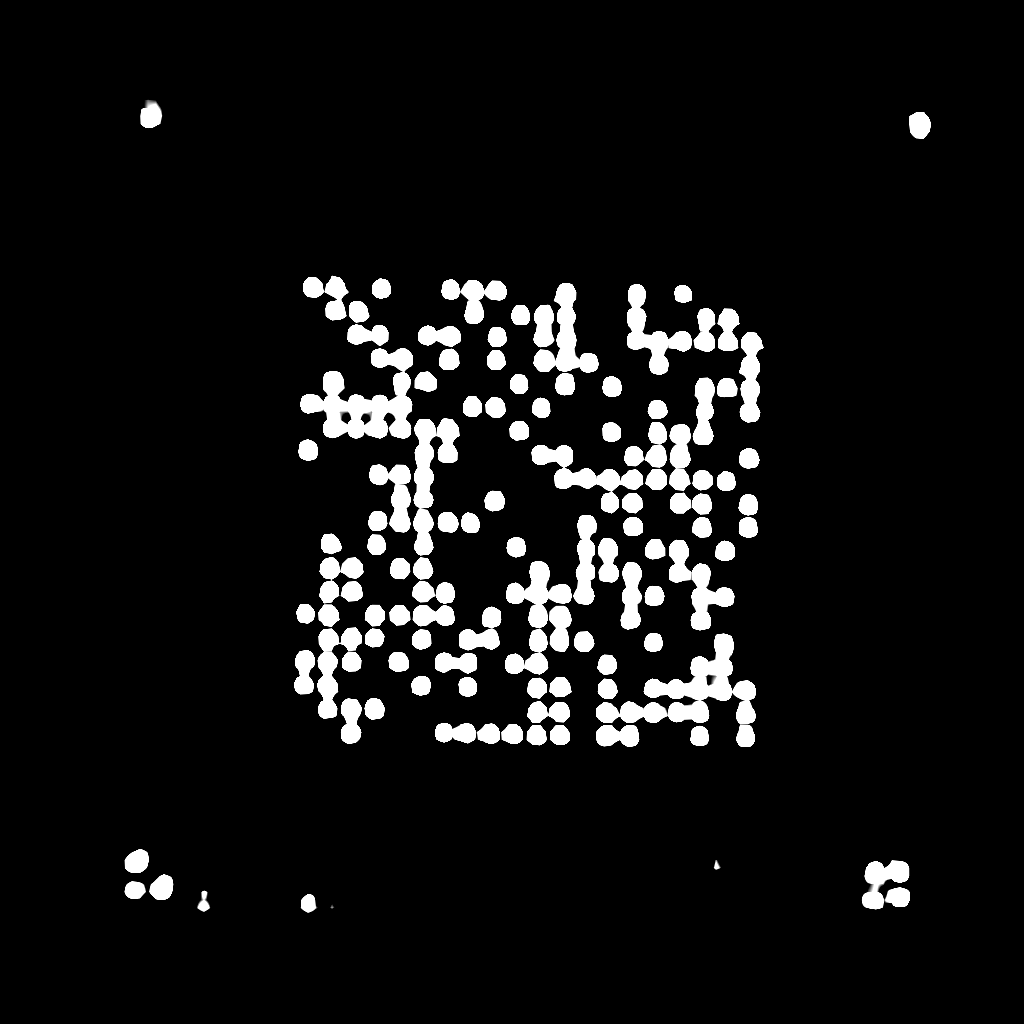}
		\caption{ResNet50 density map}
				
	\end{subfigure}
	\hfill
	\begin{subfigure}[b]{0.45\columnwidth}   
		\centering 
		\includegraphics[width=4cm, height=4cm]{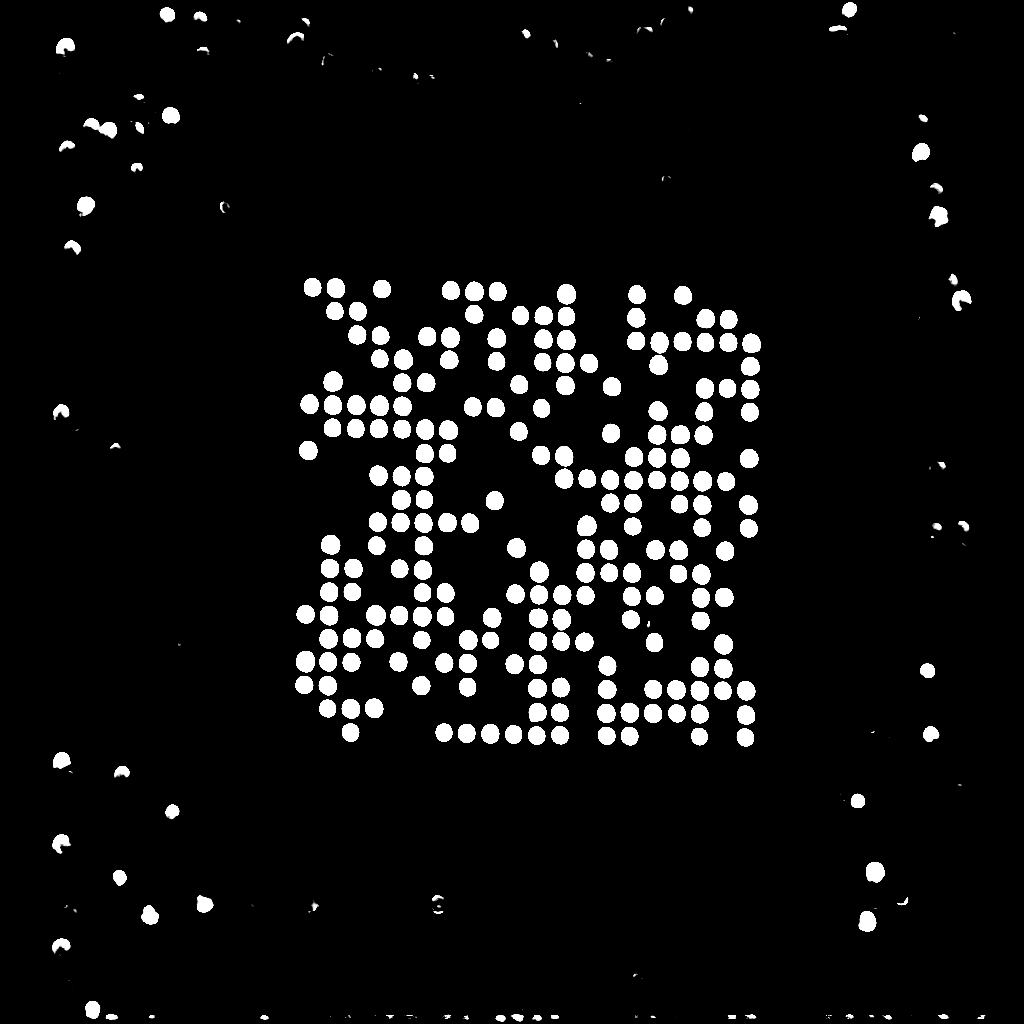}
		\caption{Unet-3DW density map}
		
	\end{subfigure}
	\caption[ ]
	{\small Density maps of Unet-Res-3DW, ResNet50 and Unet-3DW.} 
	\label{fig:eighth}-
\end{figure}

From Table~\ref{table:f1scoresNetworks}, we notice that the combined Unet-Res-3DW outperforms both of its components. We believe that the higher performance of the combined network can be partly attributed to the complementarity of the information retrieved by its two components. Figure~\ref{fig:eighth} shows the density maps computed by the three networks on a test image from the yellow 3D printed object. We notice that as a result of ResNet50 focusing on deep and abstract features, its density map contains very little noise, however, the clusters of high-valued pixels corresponding to the object's bumps tend to overlap and get connected, rather than being clearly outlined and distinct. In contrast, in the density maps from the Unet-3DW, which excels on retrieving local information, there is more noise but the high-valued pixel clusters corresponding to the object's bumps are well-localized and clearly outlined. The density map of Unet-Res-3DW seems to retain the strengths and avoid the weaknesses of its two components, as it avoids excessive noise and simultaneously produces well-localized clusters of high-valued pixels representing the object's bumps. 

Finally, we note that the performance of the three networks does not seem to drop on the white and the black 3D printed objects, even though  the texture library does not contain these two colours, and that they also cope reasonably well on the metal model, despite its very different reflectance properties. 

Table~\ref{table:results-Unet-Res-3DW} gives a more detailed description of the performance of the Unet-Res-3DW network by reporting separately the average precision and recall over the 8 training sets, as well as the performance of an ensemble algorithm determining the value of each bit of the $20 \times 20$ watermark matrix through a majority vote over 20 different images of the 3D printed object. We notice that majority vote leads to significant increases of both precision and recall, and also we note that the requirement for access to multiple images is naturally satisfied in our application scenario of watermark retrieval by a mobile phone camera. 



\begin{table}[h]
	\centering
	\caption{Average precision and recall of the Unet-Res-3DW network. Watermark retrieval from a single image, and by majority vote on ensembles of 20 images.}
	\resizebox{1\columnwidth}{!}{
	\begin{tabular}{l|cc|cc}
		\hline
		\multirow{2}{*}{Dataset} &
		\multicolumn{2}{c|}{Recall} &
		\multicolumn{2}{c}{Precision}
		 \\	 
		& {Single} & {Ensemble} & {Single} & {Ensemble} \\
		\hline
		Red & 0.82 & {\bf 0.89} & 0.76 & {\bf 0.90} \\
		yellow & 0.84 & {\bf 0.91}  & 0.88  & {\bf 0.94} \\
		green & 0.84 & {\bf 0.92} &0.88 & {\bf 0.98} \\
		white & 0.84 & {\bf 0.99} & 0.86 & {\bf 1.00} \\
		black & 0.80 & {\bf 0.92} & 0.81 & {\bf 0.91} \\
		blue & 0.81 & {\bf 0.88} & 0.80 & {\bf 0.86} \\
		purple & 0.73 & {\bf 0.83} & 0.73 & {\bf 0.84} \\
		metal & 0.74 & {\bf 0.90} & 0.83 & {\bf 0.92} \\
		\hline 
	\end{tabular}
	}
\label{table:results-Unet-Res-3DW}
\end{table}


\subsubsection{Assessing the effect of the illumination model}
\label{sec:illuminationModels}

In the study of the effect of the illumination model, each of the 8 test sets was tested against 8 network / training set combinations, the network always being the Unet-Res-3DW in ensemble mode, and the training sets corresponding to the eight illumination models in Table~\ref{table:illuminationModels}, shown in Figure~\ref{fig:illumination}. The results are summarized in Table~\ref{table:illuminationEffect}. 

\begin{table*}[h]
	\centering
	\caption{Precision and recall of Unet-Res-3DW in ensemble mode for each training/test set combination.}
	\resizebox{1\linewidth}{!}{
		\begin{tabular}{l|cc|cc|cc|cc|cc||cc|cc|cc}
			\hline
			\multirow{2}{*}{Dataset} &
			\multicolumn{2}{c|}{Illum. 1} &
			\multicolumn{2}{c|}{Illum. 2}&
			\multicolumn{2}{c|}{Illum. 3}&
			\multicolumn{2}{c|}{Illum. 4}&
			\multicolumn{2}{c||}{Illum. 5}&
			\multicolumn{2}{c|}{Illum. 6}&
			\multicolumn{2}{c|}{Illum. 7}&
			\multicolumn{2}{c}{Illum. 8}		
			\\
			& {Rec.} & {Pr.} & {Rec.} & {Pr.}& {Rec.} & {Pr.}& {Rec.} & {Pr.}& {Rec.} & {Pr.}& {Rec.} & {Pr.}& {Rec.} & {Pr.}& {Rec.} & {Pr.}\\
			\hline 
			Red & 1.00 & 1.00 & 0.81 & 0.82 & 0.90 & 0.97 & 0.96 & 0.91 & 0.92 & 0.86 & 1.00 & 1.00 & 1.00 & 1.00  & 0.88 & 0.83 \\ 
			Yellow & 1.00 & 1.00 & 0.84 & 0.82 & 0.90 & 0.89 & 0.95 & 0.89 & 0.90 & 0.88 & 0.82 & 0.90 & 1.00 & 1.00  & 0.80 & 0.85 \\ 
			Green & 0.99 & 1.00 & 0.82 & 0.83 & 0.85 & 0.81 & 0.89 & 0.84 & 0.82 & 0.80 & 0.84 & 0.98 & 0.98 & 0.95 & 0.76 & 0.70 \\ 
			White & 1.00 & 1.00 & 0.89 & 0.90 & 0.93 & 0.95 & 0.96 & 0.94 & 0.94 & 0.91 & 1.00 & 1.00 & 1.00 & 0.99 & 0.83 & 0.86 \\ 
			Black & 0.90 & 0.91 & 0.77 & 0.74 & 0.80 & 0.79 & 0.83 & 0.87 & 0.76 & 0.83 & 0.83 & 0.85 & 0.94 & 0.90 & 0.81 & 0.79 \\ 
			Blue & 0.98 & 1.00 & 0.84 & 0.88 & 0.84 & 0.86 & 0.90 & 0.90 & 0.89 & 0.84 & 0.79 & 0.95 & 0.93 & 0.88 & 0.83 & 0.75 \\ 
			Purple & 0.97 & 1.00 & 0.80 & 0.80 & 0.80 & 0.85 & 0.88 & 0.86 & 0.83 & 0.87 & 0.78 & 0.79 & 0.89 & 0.92 & 0.84 & 0.80 \\ 
			Metal & 0.96 & 0.93 & 0.82 & 0.81 & 0.79 & 0.75 & 0.91 & 0.84 & 0.78 & 0.84 & 0.84 & 0.85 & 0.91 & 0.88 & 0.73 & 0.84 \\ 
			\hline 
			Avg. & {\bf 0.98} & {\bf 0.98} & 0.82 & 0.83 & 0.85 & 0.86 & 0.91 & 0.88 & 0.86 & 0.85 & 0.86 & 0.92 & {\bf 0.96} & {\bf 0.94} & 0.81 & 0.80 \\ 
			St.d. & 0.03 & 0.04 & 0.04 & 0.05 & 0.05 & 0.08 & 0.05 & 0.04 & 0.07 & 0.03 & 0.09 & 0.08 & 0.04 & 0.05 & 0.05 & 0.05 \\ 
			\hline 
		\end{tabular}
	}\\
	\label{table:illuminationEffect}
\end{table*}

From Table~\ref{table:illuminationEffect} we see that, as expected, the choice of illumination model for generating synthetic data can have a significant effect on the performance of the network, which in some cases can exceed 15\% in both the precision and the recall rates. Interestingly, both light probes and synthetic illumination models can give good results, as we can see from the illumination models 1 and 7, which are a light probe and a synthetic model, respectively, and they trained the two best performing networks. We notice that in both cases there is a similarity between the scene corresponding to the illumination model and the environment in which the test set images were taken, that is, indoors well-lit scenes with strong ambient light coming from the windows. In contrast, the illumination model 2, which lacks ambient light, and illumination models 3 and 6, which correspond to outdoors environments, do not perform as well. 

\begin{figure}
	\centering
	\includegraphics[width=0.45\columnwidth]{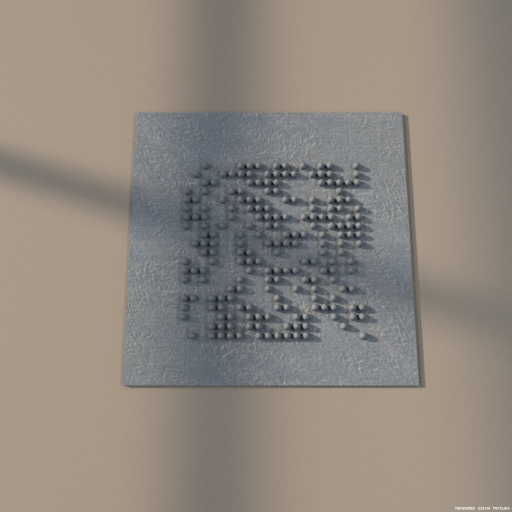} \hskip 0.5cm
	\includegraphics[width=0.45\columnwidth]{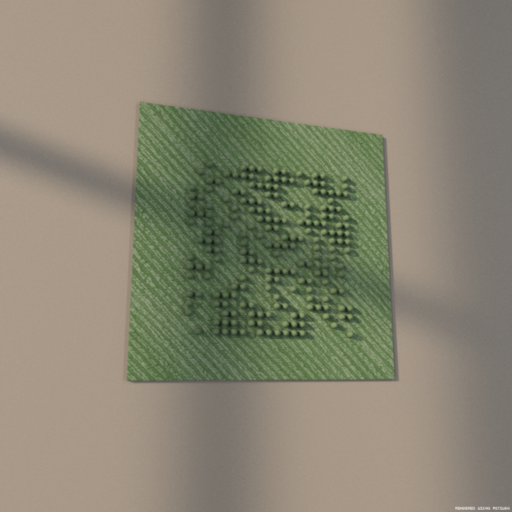} 	
	\caption{Training images generated with the illumination model 8.} 
	\label{fig:group8}
\end{figure}

The illumination models 4 and 5 do not perform well, even though they are light probes from environments similar to the test environment, that is, indoors or covered outdoors scenes with strong ambient light. However, as we can see from Table~\ref{table:illuminationModels}, their resolution is low, and the rendered images lack crucial detail. We also notice that the illumination model 8 does not perform well, despite its close resemblance to the test environment. In fact, the illumination model 8, with morning sky light entering the room through windows, is technically more faithful to the test environment than illumination model 7, where the windows are area lights, i.e., light emitting surfaces. Nevertheless, the complexity of the 3D scene in illumination model 8 means that various uncontrolled heavy shadows appear, see Figure~\ref{fig:group8}, impacting the quality of the synthetic training set.  

%% file: sec5.tex
\section{Conclusion} 
\label{section V}

We studied the effect of the illumination model on the quality of synthetic training datasets for deep neural networks. The images of the training sets are renderings of planar surfaces, with $20 \times 20$ bit arrays imprinted on them in the form semi-spherical bumps, arranged at the nodes of a regular grid. Several network/training set combinations were tested on images captured from 3D printings of these 3D models, the various training sets corresponding to different illumination models, some of them based on natural environments recorded in light probes and some others based on synthetic light modelling. Our main findings can be summarized as follows. 

Appropriate choice of illumination model can lead to increases of the accuracy rates in excess of 15\%, however, good network performance can be achieved using either captured natural light, or synthetic illumination models. In both cases, the degree of similarity between the illumination model and the natural environment in which the test images are captured is important. In the case of captured natural light, the resolution of the light probe is also an important factor. Finally, uncontrolled artifacts from complex synthetic illumination models, and in particular shadows, can degrade network performance. 

Our experiment shows that the impact of the illumination model is significant, and the potential gains from its optimization can easily exceed those from the optimization of the network architecture. However, we also note that the main findings of the experiment could have either been predicted beforehand, or could have been deduced by visual inspection of few representative images of the training set. In the future, we would like to establish relationships between the illumination model and classifier's performance that are neither obvious, nor can be deduced by visual inspection of the training set. We expect these relationships to be subtler than the ones we have studied here, and thus, a larger scale experiment would be needed to establish them.